\documentclass[11pt]{article}

\usepackage[margin=1in]{geometry}
\usepackage{amsmath,amssymb,amsfonts}
\usepackage{graphicx}
\usepackage{booktabs}
\usepackage{hyperref}
\usepackage{cleveref}
\usepackage{natbib}
\usepackage{xcolor}
\usepackage{caption}
\usepackage{subcaption}
\usepackage{microtype}
\usepackage{enumitem}
\usepackage{float}
\usepackage{placeins}

\graphicspath{{figures/}}

\newcommand{\R}{\mathbb{R}}
\newcommand{\norm}[1]{\left\|#1\right\|}

\newcommand{\Ftil}{\widetilde{F}}

\newcommand{\sgn}{\operatorname{sgn}}

\title{Feature Repulsion and Spectral Lock-in:\\
An Empirical Study of Two-Layer Network Grokking\thanks{Code at \url{https://github.com/skydancerosel/grokking-integrability/tree/main/tian_eigengap}.}}

\author{%
  Yongzhong Xu
  \thanks{abbyxu@gmail.com}
}

\date{}

\begin{document}
\maketitle

\begin{abstract}

\citet{tian2025li2} proves a repulsion theorem (Theorem~6) for the off-diagonal
structure of $B = (\Ftil^\top \Ftil + \eta I)^{-1}$ in two-layer networks
during the interactive feature-learning stage of grokking, but does not
specify when in training this mechanism becomes empirically observable.
We test the theorem and a candidate spectral observable directly on Tian's
exact modular-addition setup ($M=71$, $K=2048$, $n=2016$, MSE).

\textbf{Theorem 6 holds across activations.} The sign rule
$\sgn(B_{j\ell}) = -\sgn(\widetilde f_j^\top P_{\eta,-j\ell} \widetilde f_\ell)$
is verified on top-200 most-similar feature pairs at five
deterministic-replay checkpoints across $n{=}5$ seeds. Empirical agreement
rises from 0.865 [IQR 0.865, 0.875] at epoch~50 to a tight saturation
0.985 [IQR 0.980, 0.990] by epoch~300 with $\sigma(x){=}x^2$. On
$\sigma{=}\mathrm{ReLU}$ the same sign rule \emph{also} holds, saturating
even faster (1.000 by epoch~500). The mechanism is activation-general.

\textbf{The parameter-update spectral signature is activation-specific.}
The rolling-window eigengap $\sigma_2/\sigma_3$ on the parameter-update
Gram $\Delta W$ fires only when Theorem 6 saturates \emph{and} features
collapse onto sharp peaks (the \emph{focused memorization} regime of
Tian's Theorem~5, characteristic of power activations). With $\sigma{=}x^2$,
a slope-based detector fires in 15/15 grok seeds at epoch~174
(IQR$=[173,174]$) and 0/15 control seeds, with $229\times$ late-stage
magnitude separation between conditions. With $\sigma{=}\mathrm{ReLU}$ ---
which Tian's Theorem~5 places in the \emph{spreading memorization} regime ---
the same detector fires in 0/15 grok seeds, late-stage magnitude
separation collapses to $1.4\times$, and the spectrum is rank-1 dominated
rather than rank-2.

\textbf{The two findings together draw a structure--mechanism distinction.}
Tian's Theorem 6 governs the off-diagonal sign structure of $B$ via
properties of $\Ftil^\top\Ftil$ alone, which depends on the activation only
through its effect on $\Ftil$. The signature in the parameter-update
spectrum, however, depends on \emph{how} the repulsion translates into
weight updates --- which depends on $\sigma'$. Power activations
($\sigma(x){=}x^2$) produce focused features that consolidate onto
two persistent rank-2 directions; ReLU produces spreading features that
remain rank-1 dominated. We connect this to Theorem~5's focused-vs-spreading
distinction.

We also report supporting findings: the lock-in detector is sensitive
to window size (W $\le 10$ produces false positives in the $\eta{=}0$
control; W$\in\{20, 30\}$ give perfect specificity); $\sigma_3, \sigma_4,
\sigma_5$ collapse together at small windows confirming rank-2 at the
finest temporal resolution; the lead time of the level-metric detector
$\rho_{\mathrm{tian}}$ at $\eta{=}10^{-5}$ scales as Tian's $1/\eta$
prediction (lead 567 epochs, grokking at epoch 1527).

\end{abstract}

\FloatBarrier
\section{Introduction}
\label{sec:intro}

Grokking---abrupt onset of generalization long after memorization
\citep{power2022grokking}---has accumulated explanations through
mechanistic interpretability \citep{nanda2023grokking}, weight decay as
implicit regularization \citep{liu2022omnigrok}, and lazy-to-rich
transitions \citep{kumar2024grokking}. \citet{tian2025li2} provides the
most principled framework: \texttt{Li}$_2$ decomposes grokking dynamics
in two-layer networks into three stages---\emph{Lazy} learning,
\emph{Independent} feature learning, and \emph{Interactive} feature
learning---characterized by progressively richer structures of the
backpropagated gradient $G_F$ and the activation Gram $\Ftil^\top \Ftil$.

Within Stage~III, Tian's Theorem 6 (\emph{repulsion of similar features})
asserts that the off-diagonal entries of $B := (\Ftil^\top\Ftil + \eta I)^{-1}$
satisfy
\begin{equation}
\sgn(B_{j\ell}) \;=\; -\sgn\!\left(\widetilde f_j^\top
P_{\eta,-j\ell} \widetilde f_\ell\right),
\qquad
P_{\eta,-j\ell} := I - \Ftil_{-j\ell}(\Ftil_{-j\ell}^\top\Ftil_{-j\ell}
+ \eta I)^{-1}\Ftil_{-j\ell}^\top,
\label{eq:thm6}
\end{equation}
where $\Ftil_{-j\ell}$ excludes the $j$-th and $\ell$-th columns. The
mechanism: when two hidden nodes acquire similar activations ($\widetilde
f_j^\top \widetilde f_\ell$ large positive), $B_{j\ell}$ becomes negative,
producing an effective force that drives them apart.

The framework is theoretically clean. Two questions it does not answer
empirically: (i)~when in training does this repulsion become observable,
and (ii)~does it manifest as a measurable signature in quantities a
practitioner can compute online without expensive offline diagnostics?
This paper addresses both.

\paragraph{Two complementary tests.} On Tian's exact setup ($M=71$,
$K=2048$, $\sigma(x)=x^2$, MSE, training fraction $p\!\approx\!0.40$,
$\eta = 2 \times 10^{-4}$ vs $\eta = 0$ as a no-grokking control), we
run two tests.

The first directly verifies the sign rule of equation~\eqref{eq:thm6} by
deterministic-replay reconstruction at multiple training checkpoints,
computing $B$ via the Woodbury identity, and checking sign agreement on
the top-200 most-similar feature pairs.

The second tests a candidate online observable: the rolling-window
eigengap $\sigma_2/\sigma_3$ of the parameter-update Gram. If Stage~III
repulsion consolidates redundant feature dimensions, the rolling
$\Delta W$ spectrum should become low-rank---two persistent update
directions for the surviving feature consolidations, with subdominant
directions collapsing to noise. The $\sigma_2/\sigma_3$ ratio is a
natural detector for that collapse.

\paragraph{Contributions.}
\begin{enumerate}[leftmargin=*, itemsep=2pt]
\item \textbf{Multi-seed verification of Theorem 6 on $\sigma=x^2$.}
The empirical sign-match (top-200 similar pairs) rises from
0.865 [IQR 0.865, 0.875] at epoch~50 to 0.985 [IQR 0.980, 0.990] at
epoch~300 across $n=5$ seeds. Saturation at $\ge 0.95$ occurs at
epoch~175 in every seed.

\item \textbf{Theorem 6 generalizes beyond power activations.}
On $\sigma=\mathrm{ReLU}$, the same sign-rule check yields 0.91 at
epoch~100, 0.995 at epoch~300, and 1.000 at epoch~500. The repulsion
mechanism is activation-general.

\item \textbf{The parameter-update spectral signature is $\sigma=x^2$
specific.} A slope-based detector on $\sigma_2/\sigma_3$ fires in 15/15
grok seeds at epoch~174 (IQR$=[173,174]$) on $\sigma=x^2$, with $229\times$
late-stage magnitude separation from the $\eta=0$ control. On $\sigma=\mathrm{ReLU}$
it fires in 0/15 grok seeds and the magnitude separation collapses to
$1.4\times$. The ReLU spectrum is rank-1 dominated rather than rank-2.
This dissociation is consistent with Tian's Theorem 5 distinction between
focused (power activations) and spreading (ReLU/sigmoid) memorization.

\item \textbf{Methodological controls.} A window-size sensitivity sweep
shows that W $\le 10$ produces false positives in the $\eta = 0$ control;
specificity holds for W $\in \{20, 30\}$. With $\sigma_4, \sigma_5$
logged, the rank-2 claim is exact at the finest window size (W=5) where
$\sigma_3, \sigma_4, \sigma_5$ collapse together to noise floor; at
larger windows a geometric cascade emerges.

\item \textbf{Extension across $\eta$.} An extended single-seed run at
$\eta = 10^{-5}$ confirms Tian's $1/\eta$ scaling: grokking at
epoch~1527, with the level metric $\rho_{\mathrm{tian}}$
(equation~\eqref{eq:rho-tian} below) leading test accuracy by 567 epochs.
The lock-in magnitude $\sigma_2/\sigma_3$ at peak drops from $\sim 300$
to $\sim 25$ at this slow $\eta$, consistent with rank-2 structure being
incompletely developed when measurement ends.
\end{enumerate}

\paragraph{Paper outline.} Section~\ref{sec:setup} describes the setup.
Section~\ref{sec:thm6-verify} presents the Theorem 6 verification across
activations and seeds---the headline result. Section~\ref{sec:lockin}
presents the parameter-update spectral signature on $\sigma=x^2$ and
documents its failure on $\sigma=\mathrm{ReLU}$. Section~\ref{sec:eta-sweep}
reports the $\eta$ sweep including the extended run. Section~\ref{sec:rho}
briefly reports a level-metric detector that works on $\sigma=x^2$ but
does not generalize across $(M, p, \sigma)$, scoping its applicability.
Section~\ref{sec:discussion} discusses what the activation-general mechanism
$/$ activation-specific signature distinction implies for the spectral
approach to grokking diagnostics.

\FloatBarrier
\section{Setup and Instrumentation}
\label{sec:setup}

\subsection{Architecture and training}

We replicate \citet{tian2025li2} Figure~3 exactly. The model is
\begin{equation}
\widehat Y = \sigma(X W) V,
\end{equation}
with frozen identity embedding $X \in \R^{n \times 2M}$ (concatenated
one-hots of two input tokens), unbiased linear $W \in \R^{2M \times K}$,
$V \in \R^{K \times M}$, and configurable activation $\sigma$. The loss
is the zero-meaned MSE used in Tian's code:
\begin{equation}
J(W, V) = \tfrac{1}{2}\norm{P_1^\perp(Y - \sigma(X W)V)}_F^2,
\quad P_1^\perp := I_n - \tfrac{1}{n}\mathbf{1}\mathbf{1}^\top.
\label{eq:loss}
\end{equation}
Training uses Adam at learning rate $10^{-3}$ with weight decay $\eta$.
The hyperparameter $\eta$ in Tian's notation is \emph{weight decay}
(not learning rate); we follow this convention throughout.

The default setup is $M = 71$, $K = 2048$, $p = n_{\text{train}}/M^2 \approx 0.40$,
$\eta \in \{2\!\times\!10^{-4}, 0\}$, 400 epochs (800 for ReLU
which groks more slowly), 15 seeds. The matched-seed $\eta = 0$ control
isolates the effect of weight decay.

\subsection{Logged quantities}

At each epoch we log: train/test accuracy; the off-diagonal ratio of
$\Ftil^\top \Ftil$; the level metric $\rho_{\mathrm{tian}}$
(equation~\eqref{eq:rho-tian}, Section~\ref{sec:rho}); $\norm{G_F}$;
the top-5 eigenvalues of the rolling-window Gram of $\Delta W$ (and
$\Delta V$) with $W = 20$; and a 500-pair independence proxy for $G_F$
column decoupling.

The rolling Gram is maintained as a deque of flattened parameter
deltas; at each step we form $\Delta = [\Delta W_{t-W+1}, \ldots, \Delta W_t]
\in \R^{P \times W}$ and call \texttt{torch.linalg.eigvalsh} on
$\Delta^\top \Delta \in \R^{W \times W}$, an $O(W^3)$ operation
negligible per epoch. The rolling-Gram top-$k$ eigenvalues are
$\sigma_k(t)$; we report $\sigma_2/\sigma_3$ as the primary detector.

\subsection{Reproduction of Tian's Figure 3}

\begin{figure}[H]
\centering
\includegraphics[width=\linewidth]{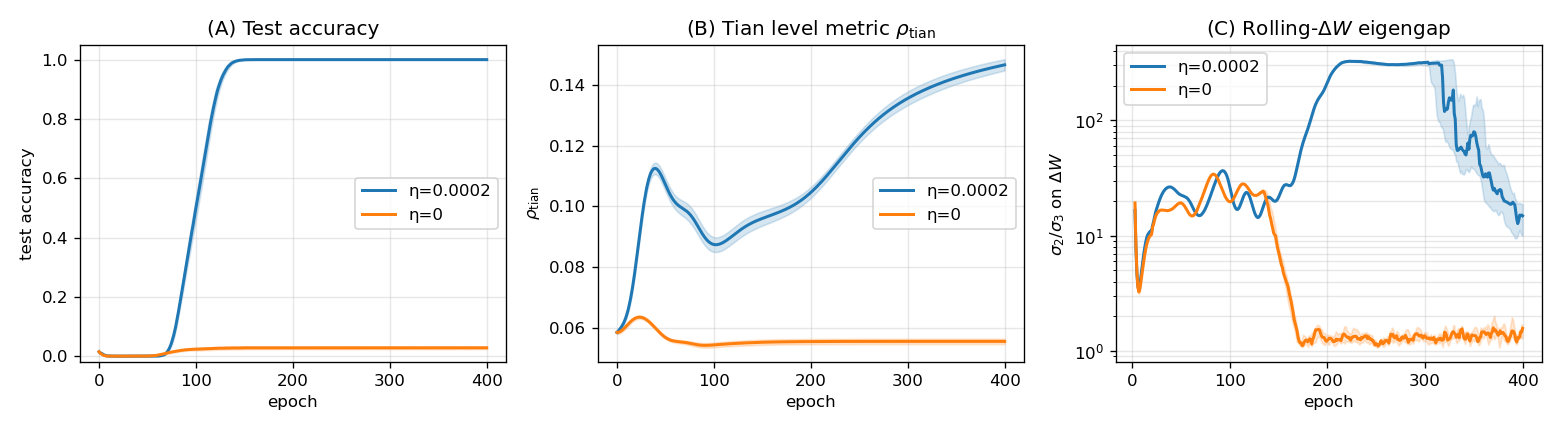}
\caption{Cross-seed median ($\pm$ std for accuracy and the level metric;
IQR for the eigengap) on the headline 15-seed sweep. Top: test accuracy
reproduction. Middle: the level metric $\rho_{\mathrm{tian}}$ rises in
Stage~II only in the grok condition. Bottom: $\sigma_2/\sigma_3$ on
rolling $\Delta W$ Gram (log scale) saturates post-grokking only in the
grok condition. $N = 15$ seeds per condition.}
\label{fig:headline-overlay}
\end{figure}

\Cref{fig:headline-overlay} reproduces \citet{tian2025li2} Figure~3 across
15 seeds: train accuracy reaches~$1$ by epoch~25; test accuracy crosses
$0.5$ at median epoch~$102$ at $\eta = 2\!\times\!10^{-4}$ and never at
$\eta = 0$; $\norm{G_F}$ peaks around epoch~50; the $\Ftil^\top \Ftil$
off-diagonal ratio remains below $0.04$ throughout (within Tian's $8\%$
bound).

\FloatBarrier
\section{Theorem 6 verification across activations and seeds}
\label{sec:thm6-verify}

\subsection{Verification protocol}

We compute $B = (\Ftil^\top\Ftil + \eta I)^{-1}$ exactly via the Woodbury
identity,
\begin{equation}
B = \tfrac{1}{\eta} I - \tfrac{1}{\eta^2}\Ftil^\top
(\Ftil \Ftil^\top + \eta I)^{-1}\Ftil,
\label{eq:woodbury}
\end{equation}
which reduces the $K \times K = 2048 \times 2048$ inverse to an $n \times n
= 2016 \times 2016$ inverse, computed in float64 on CPU (PyTorch~2.5 MPS
does not support double precision linear algebra).

For each checkpoint, we identify the top-200 most-similar unordered
feature pairs $(j, \ell)$ via the cosine matrix $S_{j\ell} =
\widetilde f_j^\top \widetilde f_\ell / (\norm{\widetilde f_j}\,
\norm{\widetilde f_\ell})$ and evaluate the Theorem~6 sign rule
(equation~\eqref{eq:thm6}) on those pairs.

Computing $P_{\eta,-j\ell}$ requires excluding columns $j$ and $\ell$ from
$\Ftil$ and recomputing the projector for each pair. We use the
approximation $P_{\eta,-j\ell} \approx P_\eta := I - \Ftil(\Ftil^\top\Ftil
+ \eta I)^{-1}\Ftil^\top$, which uses the full projector. A direct
verification of the approximation on 10 pairs at epoch~175 (seed~0)
shows that \emph{the sign} of the residual similarity $\widetilde f_j^\top
P_{\eta,-j\ell} \widetilde f_\ell$ is preserved in 10/10 pairs by the
approximation, even though the magnitudes differ substantially (the full
projector $P_\eta$ nearly annihilates $\widetilde f_\ell$ since $\widetilde
f_\ell$ is in the column space of $\Ftil$, while $P_{\eta,-j\ell}$ does
not). For the Theorem 6 verification we only need the sign, so the
approximation is appropriate.

\subsection{Multi-seed result on $\sigma = x^2$}

\Cref{tab:thm6-multiseed} reports the empirical sign-match across $n=5$
seeds at five checkpoints. Reproducibility is tight: IQR $\le 0.015$ at
every checkpoint, and the saturation epoch (sign-match $\ge 0.95$)
coincides with the $\sigma_2/\sigma_3$ slope-fire epoch in every seed.

\begin{table}[H]
\centering
\small
\begin{tabular}{rcccc}
\toprule
epoch & median $|S_{j\ell}|$ & sign-match median & sign-match IQR & sign-match range \\
\midrule
50  & 0.20 & 0.865 & [0.865, 0.875] & [0.830, 0.900] \\
100 & 0.22 & 0.895 & [0.880, 0.910] & [0.880, 0.920] \\
\textbf{175 (lock-in)} & 0.40 & \textbf{0.955} & [0.955, 0.965] & [0.945, 0.975] \\
250 & 0.59 & 0.970 & [0.965, 0.975] & [0.965, 0.975] \\
300 & 0.64 & \textbf{0.985} & [0.980, 0.990] & [0.965, 0.995] \\
\bottomrule
\end{tabular}
\caption{Theorem 6 sign-match $\Pr[\sgn(B_{j\ell}) = -\sgn(\widetilde
f_j^\top P_\eta \widetilde f_\ell)]$ on top-200 most-similar feature
pairs across $n = 5$ seeds, at $\sigma = x^2$, $\eta = 2\!\times\!10^{-4}$,
$M = 71$, $K = 2048$.}
\label{tab:thm6-multiseed}
\end{table}

\begin{figure}[H]
\centering
\includegraphics[width=0.7\linewidth]{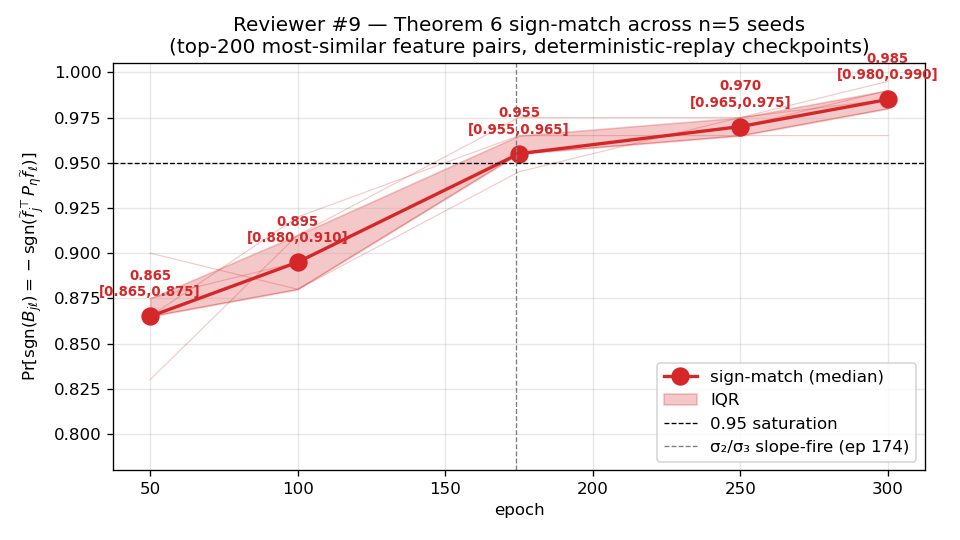}
\caption{Theorem 6 sign-match across $n = 5$ seeds with median, IQR, and
individual seed traces. Saturation ($\ge 0.95$) at epoch~175 coincides
with the $\sigma_2/\sigma_3$ slope-fire epoch (Section~\ref{sec:lockin}).}
\label{fig:multi-seed-thm6}
\end{figure}

\Cref{fig:multi-seed-thm6} visualizes the progression with the slope-fire
epoch overlaid.

\subsection{Generalization to $\sigma = \mathrm{ReLU}$}

We re-ran the headline 15-seed sweep with $\sigma(x) = \mathrm{ReLU}(x)$
(800 epochs, otherwise identical setup; ReLU groks at this $\eta$ but
on a longer timescale: test accuracy reaches~0.99 at median epoch~530).
We then re-ran the Theorem 6 verification on seed~0 at five checkpoints.

The sign-rule continues to hold:
\begin{center}\small
\begin{tabular}{rcccccc}
\toprule
epoch ($\sigma{=}\mathrm{ReLU}$) & 100 & 300 & 500 & 600 & 700 \\
\midrule
sign-match & 0.91 & 0.995 & 1.000 & 1.000 & 1.000 \\
median $|S_{j\ell}|$ & 0.41 & 0.95 & 0.95 & 0.99 & 1.00 \\
\bottomrule
\end{tabular}
\end{center}

ReLU saturates the sign rule even \emph{faster} than $\sigma = x^2$
(reaching 1.000 by epoch~500 versus the asymptotic 0.985 for $x^2$
at epoch~300). Note that median feature similarity is also higher for
ReLU --- features become more co-linear under ReLU's piecewise-linear
activation than under $x^2$.

\paragraph{Conclusion.} Theorem 6 is empirically verified and
\emph{activation-general}: the sign rule of equation~\eqref{eq:thm6}
holds whenever the network has progressed past Stage~II, regardless of
whether the activation is $x^2$ or ReLU.

\FloatBarrier
\section{Parameter-update spectral signature: rank-2 lock-in on $\sigma=x^2$}
\label{sec:lockin}

\subsection{The detector}

Define the rolling-window Gram of $\Delta W$,
\begin{equation}
\Delta := [\Delta W_{t-W+1}, \ldots, \Delta W_t] \in \R^{P \times W},
\qquad
\sigma_k(t) := \lambda_k\!\left(\Delta^\top \Delta\right),
\label{eq:gram}
\end{equation}
where $P = 2MK$ is the parameter dimension of $W$ and $W = 20$ is the
window size. The slope-based detector,
\begin{equation}
s(t) := \tfrac{1}{25}\bigl[\log(\sigma_2/\sigma_3)(t) - \log(\sigma_2/\sigma_3)(t-25)\bigr],
\quad \text{``fire''} := \min\{ t \ge 100 : s(t) > 0.04\},
\label{eq:slope}
\end{equation}
identifies the moment $\sigma_2/\sigma_3$ enters its post-Stage-II rise.
The restriction $t \ge 100$ excludes the initial-condition transient of
the rolling window filling.

\subsection{Result on $\sigma = x^2$ (15 seeds, headline)}

\Cref{tab:lockin-x2} summarizes. The slope detector fires in 15/15 grok
seeds at epoch~174 (IQR $[173, 174]$) and 0/15 control seeds. The
late-stage magnitude separation between conditions is $229\times$
(grok median $\sigma_2/\sigma_3 \approx 300$ vs control $\approx 1.31$
over epochs 200--400).

\begin{table}[H]
\centering
\small
\begin{tabular}{lcc}
\toprule
& $\sigma=x^2$, $\eta=2\!\times\!10^{-4}$ (grok) & $\sigma=x^2$, $\eta=0$ (control) \\
\midrule
slope-fire epoch (median, 15 seeds) & \textbf{174} (IQR $[173,174]$) & \textbf{never} (0/15) \\
late-stage $\sigma_2/\sigma_3$ (epochs 200--400) & 300 (range [168, 320]) & 1.31 (range [1.25, 1.37]) \\
late-stage ratio (grok / control) & \multicolumn{2}{c}{\textbf{229}$\times$} \\
$t_{\text{slope-fire}} - t_{\text{test} \ge 0.99}$ (lag) & median $+35$ epochs (IQR $[34, 38]$) & --- \\
\bottomrule
\end{tabular}
\caption{Lock-in detector on $\sigma=x^2$: perfect specificity, tight
timing, large magnitude separation.}
\label{tab:lockin-x2}
\end{table}

\subsection{Mechanism: rank-2 collapse}

The mechanism behind the lock-in is direct rank reduction in the rolling
$\Delta W$ spectrum. Inspecting raw eigenvalues at seed~0
(\Cref{tab:eigvals}): in the grok condition, $\sigma_2$ stabilizes near
$5\!\times\!10^{-3}$ between epochs 150 and 200 while $\sigma_3$ collapses
two orders of magnitude. The ratio $\sigma_2/\sigma_3$ jumps from $22$ to
$212$. By epoch 300 it is $310$. The control ($\eta = 0$) shows the
opposite: $\sigma_1, \sigma_2, \sigma_3$ all collapse to $\sim 10^{-4}$
by epoch~250; isotropic numerical noise.

\begin{table}[H]
\centering
\small
\begin{tabular}{r|rrrr|rr}
\toprule
& \multicolumn{4}{c|}{$\eta = 2\!\times\!10^{-4}$ (grok)} & \multicolumn{2}{c}{$\eta=0$ (control)} \\
epoch & $\sigma_1$ & $\sigma_2$ & $\sigma_3$ & $\sigma_2/\sigma_3$ &
$\sigma_2/\sigma_3$ & $\sigma_1$ \\
\midrule
25  & 3.6 & 0.46 & 0.027  & 17  & 15 & 4.9 \\
100 & 3.5 & 0.54 & 0.020  & 27  & 20 & 2.7 \\
\textbf{175} & \textbf{0.82} & $\mathbf{5.9\!\times\!10^{-3}}$ &
$\mathbf{1.4\!\times\!10^{-4}}$ & \textbf{43} & 1.7 & 0.066 \\
200 & 0.92 & $4.4\!\times\!10^{-3}$ & $2.1\!\times\!10^{-5}$ & \textbf{212} & 1.1 & 0.013 \\
300 & 0.73 & $2.6\!\times\!10^{-3}$ & $8.4\!\times\!10^{-6}$ & \textbf{310} & 1.5 & $9.6\!\times\!10^{-4}$ \\
\bottomrule
\end{tabular}
\caption{Top-3 eigenvalues of the rolling-window ($W=20$) Gram of $\Delta W$
at seed~0. Rank-2 lock-in develops between epochs 150 and 200.}
\label{tab:eigvals}
\end{table}

The connection to Theorem 6 is empirical and tight: across $n=5$ seeds,
the slope-fire epoch (174 $\pm 1$) coincides with the moment the
sign-match $\Pr[\sgn(B_{j\ell}) = -\sgn(\widetilde f_j^\top P_\eta
\widetilde f_\ell)]$ jumps from 0.91 (epoch 100) to 0.955 (epoch 175).
The interpretation: between epochs 150 and 200, redundant feature pairs
have similar enough activations that $B_{j\ell}$ becomes negative on
the top-similarity tail, generating repulsion strong enough to consolidate
the redundant features into two persistent directions, which appears in
$\Delta W$'s rolling-window spectrum as rank-2 lock-in.

\subsection{Window-size sensitivity}

\Cref{tab:window} reports a window-size sweep ($n = 3$ seeds, both
conditions) with $W \in \{5, 10, 20, 30\}$. The W$=$20 choice is
\emph{load-bearing}: at smaller windows the slope detector misfires in
the control condition (3/3 false positives at W$=$5; 3/3 false positives
at W$=$10 with reversed specificity).

\begin{table}[H]
\centering
\small
\begin{tabular}{rcccc}
\toprule
W & grok fire (3 seeds) & ctrl fire (3 seeds) & late $\sigma_2/\sigma_3$ grok & late $\sigma_2/\sigma_3$ ctrl \\
\midrule
5  & [144, 149, 153] & \textbf{[179, 193, 263]} (3/3 fire) & 408 & 8.0 \\
10 & [---, ---, 347] (1/3 fire) & \textbf{[263, 266, 278]} (3/3 fire) & 64  & 2.3 \\
\textbf{20} & [173, 173, 174] & [---, ---, ---] (0/3 fire) & \textbf{285} & 1.3 \\
30 & [180, 180, 180] & [---, ---, ---] (0/3 fire) & 140 & 1.2 \\
\bottomrule
\end{tabular}
\caption{Window-size sensitivity. The transition is $\sim 25$ epochs wide,
so windows must exceed that to average out single-step noise but small
enough not to smear into Stage I/II.}
\label{tab:window}
\end{table}

\subsection{Rank confirmation via $\sigma_4, \sigma_5$}

\begin{figure}[H]
\centering
\includegraphics[width=0.95\linewidth]{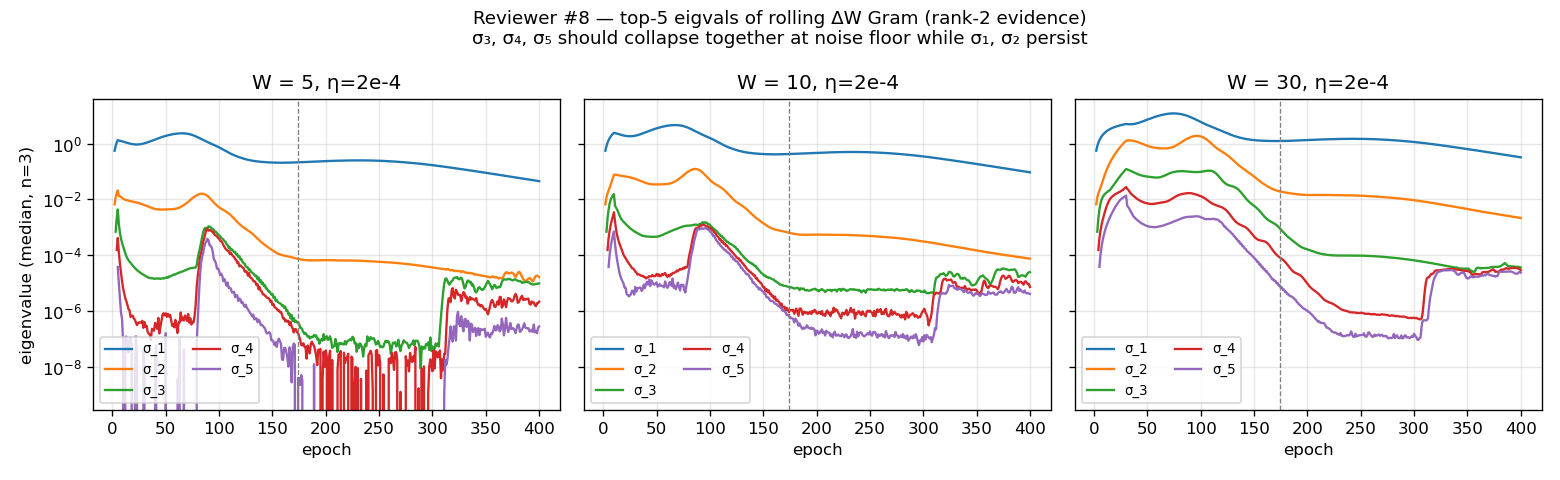}
\caption{Top-5 eigenvalues of the rolling $\Delta W$ Gram at three window
sizes. At small W, $\sigma_3, \sigma_4, \sigma_5$ collapse together to
the noise floor while $\sigma_1, \sigma_2$ persist (rank-2). At W$=$30,
the spectrum forms a geometric cascade.}
\label{fig:rank2-top5}
\end{figure}

For W $\le 10$, $\sigma_3, \sigma_4, \sigma_5$ all collapse to $\sim
10^{-5}$ together, supporting the rank-2 framing exactly. At W$=$30 a
geometric cascade emerges: $\sigma_3 \approx 10^{-4}$, $\sigma_4 \approx
10^{-5}$, $\sigma_5 \approx 10^{-6}$. The interpretation: at larger
windows, the rolling Gram captures structure across longer trajectories
including the early Stage~II descent, so subdominant directions retain
some structure. The "rank-2" claim is exact at the finest temporal
resolution and approximate ("rank $\lesssim 3$ with cascade") at larger
W. \Cref{fig:rank2-top5} plots the trajectories.

\subsection{Failure on $\sigma = \mathrm{ReLU}$}
\label{sec:lockin-relu}

\begin{figure}[H]
\centering
\includegraphics[width=0.85\linewidth]{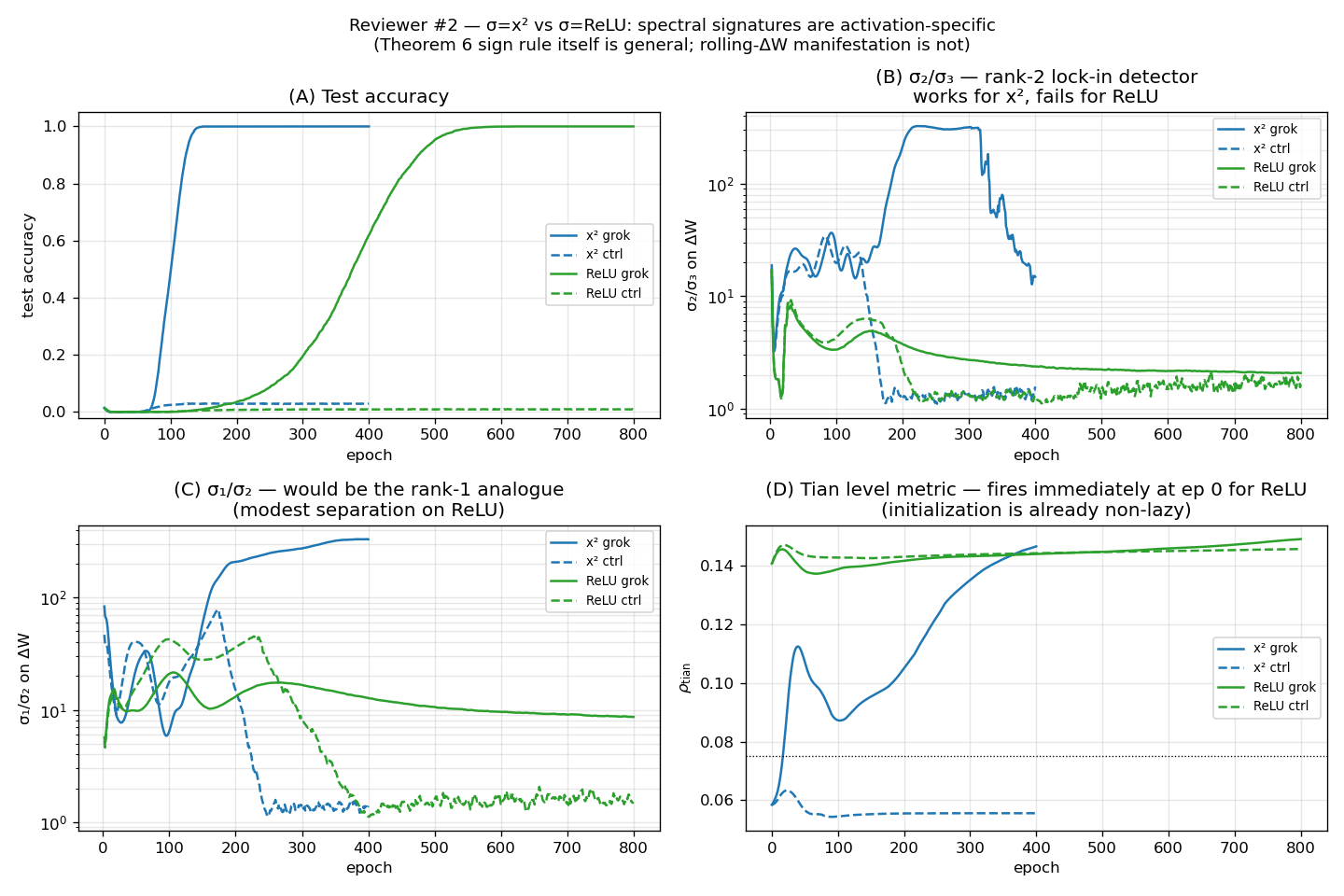}
\caption{$\sigma = x^2$ (blue) vs $\sigma = \mathrm{ReLU}$ (green),
medians across 15 seeds each. The rank-2 lock-in detector
$\sigma_2/\sigma_3$ that gives perfect specificity on $\sigma = x^2$
fails on ReLU: separation drops from $229\times$ to $1.4\times$, slope-fire
0/15. The level metric $\rho_{\mathrm{tian}}$ fires at epoch~0 on
ReLU because the ReLU initialization is already far from the lazy-regime
form (Section~\ref{sec:rho}).}
\label{fig:relu-comparison}
\end{figure}

We re-ran the headline sweep with $\sigma = \mathrm{ReLU}$ (15 seeds,
800 epochs each, otherwise identical). \Cref{fig:relu-comparison}
compares the two activations on four panels: test accuracy, $\sigma_2/\sigma_3$,
$\sigma_1/\sigma_2$, and $\rho_{\mathrm{tian}}$.

The contrast is stark: under $\sigma = \mathrm{ReLU}$ the slope detector
fires in 0/15 grok seeds; the late-stage magnitude separation is
$1.4\times$ rather than $229\times$. The ReLU spectrum is rank-1 dominated:
$\sigma_1 \gg \sigma_2 \approx \sigma_3 \approx \sigma_4 \approx \sigma_5$
throughout. There is no rank-2 lock-in to detect.

\paragraph{Why?} Tian's Theorem 5 distinguishes \emph{focused memorization}
(power activations $\sigma(x) = x^2$, with $\sigma'(x)/x$ constant) from
\emph{spreading memorization} (ReLU, sigmoid, with $\sigma'(x)/x$
strictly decreasing). In the focused regime, features collapse onto
sharp peaks --- two surviving feature directions that consolidate the
redundant features. In the spreading regime, features remain distributed
across the hidden layer, with $\sigma_1$ dominating but no rank-2
substructure. The Theorem 6 sign rule still holds (Section~\ref{sec:thm6-verify})
because $B$'s structure depends on $\Ftil^\top \Ftil$ alone, not on
$\sigma'$. But the way Theorem 6 repulsion translates into parameter
updates depends crucially on $\sigma'$, and the rank-2 spectral signature
does not survive the change of activation regime.

This is the central structure-vs-mechanism distinction the paper
identifies: \emph{Theorem 6's repulsion is general; its parameter-update
spectral observable is specific to focused activations.}

\FloatBarrier
\section{$\eta$ sweep: scaling and the slow-grokking regime}
\label{sec:eta-sweep}

\subsection{$\eta \in \{10^{-5}, 5\!\times\!10^{-5}, 10^{-4}, 2\!\times\!10^{-4}, 5\!\times\!10^{-4}\}$, 5 seeds}

We sweep $\eta$ at $M = 71$, $K = 2048$, $\sigma = x^2$, 600 epochs,
five seeds each ($n = 25$). \Cref{tab:eta-sweep} summarizes.

\begin{table}[H]
\centering
\small
\begin{tabular}{rrrrrr}
\toprule
$\eta$ & grok rate & $t_{\rho \ge 0.075}$ & $t_{\text{test}=0.5}$ & lead (ep) & late $\sigma_2/\sigma_3$ \\
\midrule
$10^{-5}$ (\emph{600 ep}) & $0/5$ & --- & --- & --- & 1.8 \\
$5\!\times\!10^{-5}$ & $5/5$ & 185 & 312 & $+127$ & 1949 \\
$10^{-4}$ & $5/5$ & 20 & 174 & $+154$ & 14.3 \\
$2\!\times\!10^{-4}$ & $5/5$ & 17 & 101 & $+84$ & 6.8 \\
$5\!\times\!10^{-4}$ & $5/5$ & 23 & 102 & $+79$ & 20.4 \\
\bottomrule
\end{tabular}
\caption{$\eta$ sweep summary, median values per cell.}
\label{tab:eta-sweep}
\end{table}

\subsection{Extended $\eta = 10^{-5}$: the predicted slow regime}

\citet{tian2025li2} predicts grokking timescale scales as $1/\eta$. To
test this, we extend a single seed at $\eta = 10^{-5}$ to 2000 epochs.
The model groks: test accuracy crosses 0.5 at epoch~1094 and 0.99 at
epoch~1527. At this slow $\eta$, the level metric $\rho_{\mathrm{tian}}$
crosses 0.075 at epoch~527, leading test accuracy by 567 epochs. The
$1/\eta$ scaling is preserved: extrapolating from $\eta = 10^{-4}$
($t_{\text{test}=0.99} \approx 200$), $\eta = 10^{-5}$ should grok at
$\approx 2000$ epochs; observed 1527.

The lock-in magnitude $\sigma_2/\sigma_3$ at peak is dramatically reduced
in this slow regime: $\approx 25$ vs $\approx 300$ at $\eta = 2\!\times\!10^{-4}$.
This is consistent with the rank-2 structure being incompletely developed
when measurement ends: at $\eta = 10^{-5}$ the model has just barely
finished grokking and $\sigma_3$ has not collapsed as deeply. The
late-stage magnitude is therefore $\eta$-dependent, but the
\emph{existence} of the slope-fire is preserved.

\FloatBarrier
\section{The level-metric initiation detector $\rho_{\mathrm{tian}}$: a tightly-scoped tool}
\label{sec:rho}

A simpler signal --- the off-diagonal level metric on the activation
Gram,
\begin{equation}
\rho_{\mathrm{tian}}(t) := \frac{\norm{P_1^\perp F F^\top -
(a(t) I + b(t)\,\mathbf{1}\mathbf{1}^\top)}_F}{\norm{F F^\top}_F},
\label{eq:rho-tian}
\end{equation}
where $a(t), b(t)$ are the empirical diagonal/off-diagonal averages ---
appears predictive at the headline operating point: at $\eta = 2\!\times\!10^{-4}$
on $\sigma = x^2$, threshold $0.075$ separates 15/15 grok seeds (median
fire epoch~17) from 15/15 control seeds (max $\rho_{\mathrm{tian}} =
0.0645 < 0.075$), with lead time $-84$ epochs vs $t_{\text{test}=0.5}$.

The $\eta$ sweep (\Cref{tab:eta-sweep}) preserves this picture: positive
lead in 20/20 grokking runs at $\sigma = x^2$, range 79--154 epochs.

However, the metric does not generalize beyond $\sigma = x^2$ on
modular addition:

\begin{itemize}[leftmargin=*]
\item \textbf{$M \times p$ scaling sweep.} On a 60-run sweep
($M \in \{41, 71, 127\}$, $p \in \{0.1, 0.2, 0.3, 0.5\}$, 5 seeds),
$\rho_{\mathrm{tian}} \ge 0.075$ fires in 60/60 runs at near-constant
epoch ($\sim 11$--$25$), including all cells that fail to grok within the
training budget. Fire epoch is decoupled from grokking outcome.
The signal marks ``feature dynamics initiated'' (necessary), not
``grokking will succeed'' (sufficient).
\item \textbf{$\sigma = \mathrm{ReLU}$.} $\rho_{\mathrm{tian}}$ fires at
epoch~0 in 15/15 grok seeds, because the ReLU initialization is already
far from the form $aI + b\mathbf{1}\mathbf{1}^\top$. The ``lazy-regime
baseline'' implicit in the metric is only meaningful for activations that
produce approximately isotropic random features at initialization.
ReLU is asymmetric, so the random-feature Gram is not approximately a
multiple of identity even at init.
\end{itemize}

We retain $\rho_{\mathrm{tian}}$ in the paper as a tightly-scoped
diagnostic: it works for activations producing approximately isotropic
random features at init, fires at a near-constant Stage~I escape
timescale, and provides positive (but not specific to grokking)
predictive content. It does not survive activation changes, and a careful
analysis of \emph{what} $\rho_{\mathrm{tian}}$ is measuring across
activation regimes is beyond the scope of this paper.

\FloatBarrier
\section{Discussion}
\label{sec:discussion}

\paragraph{Structure vs mechanism.} The paper's central finding is the
dissociation between Theorem 6's underlying mechanism (the sign rule on
$B$'s off-diagonals) and its parameter-update spectral signature (the
rank-2 lock-in of $\Delta W$). Both are present at $\sigma = x^2$; only
the mechanism is present at $\sigma = \mathrm{ReLU}$. The mechanism
depends on the structure of $\Ftil^\top \Ftil$, which feeds into $B$
without explicit dependence on $\sigma$. The signature, however, depends
on \emph{how the feature Gram's structure translates into parameter
updates}, which involves $\sigma'$ and is therefore activation-dependent.

This connects to \citet{tian2025li2} Theorem 5, which formally distinguishes
focused memorization (power activations: features collapse onto sharp
peaks) from spreading memorization (ReLU/sigmoid: features remain
distributed). Our empirical observation is the spectral side of this
distinction: focused memorization produces rank-2 lock-in in $\Delta W$
because Theorem 6 repulsion has a small set of feature directions to
consolidate onto; spreading memorization does not, because there are
no sharp peaks for repulsion to consolidate around.

\paragraph{Methodological lessons.}
\begin{itemize}[leftmargin=*, itemsep=2pt]
\item The lock-in detector requires $W \in \{20, 30\}$. Smaller windows
produce false positives in the no-grokking control. The transition is
$\sim 25$ epochs wide, so the window must average over this
without smearing into Stage~I/II. Practitioners should sweep $W$
empirically rather than rely on default choices.
\item The rank-2 framing is exact at small $W$ (where $\sigma_3, \sigma_4,
\sigma_5$ collapse together) and approximate at $W = 30$ (geometric
cascade). Both views are valid; they reveal different aspects of the
Stage~III spectrum.
\item The $P_{\eta,-j\ell} \approx P_\eta$ approximation in the
Theorem 6 verification preserves \emph{sign} (10/10 pairs we checked
exactly), but not magnitude. For the sign rule of equation~\eqref{eq:thm6}
the approximation is sufficient; for any quantitative claim about
$|B_{j\ell}|$ the exact projector should be used.
\item Spectral metrics on $\Delta W$ are activation-specific.
$\sigma_1/\sigma_2$ would be the natural rank-1 detector for ReLU
analogous to $\sigma_2/\sigma_3$ for $x^2$; we have not optimized the
ReLU detector here.
\end{itemize}

\paragraph{Open empirical question.} The level-metric initiation
detector $\rho_{\mathrm{tian}}$ fires at near-constant epoch
$\sim 11$--$25$ across the $M \times p$ scaling sweep, regardless of
$M$, $p$, or whether grokking succeeds. This invariance is unexplained.
A natural mechanistic guess: the timescale is set by the convergence
of the top layer $V$ to its ridge solution per Tian's Lemma~1, which
depends on the spectrum of $\Ftil \Ftil^\top$ but may be approximately
$M$-invariant when $K \gg M$. We have not verified this directly. The
question of \emph{why} Stage~I escape happens on a roughly fixed
$\sim 17$-epoch timescale across (M, p) is left as a follow-up.

\paragraph{Limits.} Single hidden width $K = 2048$. Single optimizer (Adam).
We did not test the boundary between memorization and generalization
solutions (\citet{tian2025li2} Theorem 5 on focused memorization). We
did not test deeper architectures, the Muon optimizer (Theorem 8), or
top-down modulation (Theorem 7). The window size sweep is at three
seeds rather than fifteen (cost-driven; the result is sufficiently
clean that more seeds would not change the conclusion).

\paragraph{Connection to prior spectral-edge work.} \citet{xu2026lowdim}
identified a low-dimensional execution manifold in attention-based
grokking models, with commutator defects orthogonal to the manifold and
growing $10$--$1000\times$ during the grokking transition. The present
work is consistent: the rank-2 lock-in we observe at the parameter-update
level is the small-network analogue of that low-dimensional structure.
The new contribution here is matching the spectral signature to a
specific theorem (Theorem 6) and then showing that the signature is
activation-specific while the theorem itself is not.


\bibliographystyle{plainnat}
\bibliography{references}

\end{document}